# Original Paper

# Overview and Comparison of AVS Point Cloud Compression Standard

Wei Gao[1,2], Wenxu Gao[1,2], Xingming Mu[1], Changhao Peng[1] and Ge Li[1*]

[1] *Guangdong Provincial Key Laboratory of Ultra High Definition Immersive Media Technology, School of Electronic and Computer Engineering, Peking University, Shenzhen 518055, China.*
[2] *Peng Cheng Laboratory, Shenzhen 518066, China.*

## ABSTRACT

Point cloud is a prevalent 3D data representation format with significant application values in immersive media, autonomous driving, digital heritage protection, etc. However, the large data size of point clouds poses challenges to transmission and storage, which influences the wide deployments. Therefore, point cloud compression plays a crucial role in practical applications for both human and machine perception optimization. To this end, the Moving Picture Experts Group (MPEG) has established two standards for point cloud compression, including Geometry-based Point Cloud Compression (G-PCC) and Video-based Point Cloud Compression (V-PCC). In the meantime, the Audio Video coding Standard (AVS) Workgroup of China also have launched and completed the development for its first generation point cloud compression standard, namely AVS PCC. This new standardization effort has adopted many new coding tools and techniques, which are different from the other counterpart standards. This paper reviews the AVS PCC standard from two perspectives, i.e., the related technologies and performance comparisons.

*Corresponding author: Ge Li, geli@pku.edu.cn.

## 1  Introduction

Point clouds [44, 30] and polygonal grids are commonly employed for representing 3D data [95, 91, 76, 39, 19, 110]. In comparison to polygonal grids, point clouds, consisting of a collection of data points in a coordinate system, provide a high degree of accuracy and flexibility in representing complex 3D structures [62]. Unlike polygonal grids, which are composed of vertices, edges, and faces, point clouds require no explicit connectivity information between points, simplifying the modeling process and allowing for more straightforward acquisition from LiDAR scanners, depth sensors, and other 3D scanning technologies [55, 14, 120, 28], which makes them crucial in the field of 3D data representation. Point clouds has found extensive applications in the multimedia domain [115, 29], transforming the way we engage with digital content. From the seamless integration in virtual reality environments for gaming and education to the enhancement of visual effects in films [99, 20, 109], point clouds provide a rich and immersive experience. They also play a crucial role in 3D modeling and animation, enabling the creation of realistic characters and environments. Furthermore, point clouds are instrumental in the field of augmented reality, overlaying digital information onto the physical world, thus bridging the gap between digital and real-time interactions [84]. This multifaceted utilization underscores the significance of point cloud technologies in enriching multimedia experiences, and thus the research on point cloud processing and analysis has become very popular [101, 70, 58, 118, 56, 53].

A point cloud, an assemblage of discrete data points scattered across a 3D spatial expanse [63], transcends the conventional boundaries of 2D imaging by offering a more nuanced and multifaceted representation of the physical world. Each point within this cloud is endowed with a unique set of coordinates, anchoring it within the 3D space, and carries its own attribute information, which can extend beyond basic spatial coordinates to include color, intensity, and even reflectance values [129, 9]. Unlike the structured grid of pixels in a 2D image, point clouds consist of an unordered array of points, which can be densely packed in areas of high interest while being more sparse in others. This irregularity in data distribution provides a level of flexibility and scalability that is well-suited for representing intricate geometries and dynamic scenes. The inclusion of additional attributes, such as reflectance, enriches the point cloud's descriptive power, enabling it to convey not just the visual appearance but also the material properties of the surfaces it represents. Moreover, point clouds are not limited to static representations; they can be dynamically generated and updated in real-time by various 3D scanning technologies [38], making them ideal for applications that require live environmental interaction and feedback, such as robotics navigation and augmented reality. As a result, point clouds have become an indispensable tool in a myriad of fields, from architectural scanning and urban planning to medical imaging and advanced manufacturing, where the need for accurate 3D data representation is paramount [74, 69, 25, 121, 110].

Point cloud compression [29] can be categorized into geometry compression and attribute compression based on the type of data being compressed [10, 68, 34, 67]. Geometry compression of point clouds [80, 128, 106, 73, 113, 108, 88, 104] focuses on efficiently encoding the spatial coordinates of the data points while minimizing the loss of information. This process involves techniques such as quantization, octree partitioning [85, 77], and predictive coding [42] that exploit the spatial redundancy and geometrical patterns within the point cloud to achieve high compression ratios. The goal is to represent the 3D structure with accuracy, ensuring that the reconstructed point cloud maintains the essential characteristics of the original dataset. Attribute compression [71, 81, 54, 87, 86, 22, 127] addresses the encoding of non-spatial information associated with each point in the cloud, such as color, intensity, and reflectivity. This form of compression leverages the statistical dependencies and correlations between attribute values to reduce the bit rate required for representation. Techniques like differential coding, transform coding, and predictive techniques [37] can be applied to compactly represent the attributes that enrich the visual and analytical utility of the point cloud data. The challenge lies in preserving the perceptual quality [51, 93, 92, 21, 52] and information content after compression.

The practical application of point cloud in industry faces significant challenges due to the large volume of data involved, particularly during the transmission of data from local to server. Therefore, efficient compression techniques for point cloud data have become crucial for the development of related applications [102, 103, 107, 111, 1, 26, 105, 116]. In this regard, Moving Picture Experts Group (MPEG) [15, 78], a renowned international standard group in video coding, has proposed two distinct technical approaches for point cloud coding, including Geometric-based Point Cloud Compression (G-PCC) [48, 50] and Video-based Point Cloud Compression (V-PCC) [34, 49, 50, 31, 114]. MPEG G-PCC eliminates redundancies within the point cloud data by leveraging its geometric characteristics, while MPEG V-PCC utilizes mature video coding technology to project 3D point clouds onto a 2D space. Additionally, the Audio Video coding Standard (AVS) Workgroup of China introduces AVS Point Cloud Compression (PCC) standard [46, 50], which is a well-known audio and video coding standardization organization.

The timeline of the development of AVS PCC standard is depicted in Figure 1 [7, 46]. In March 2019, the inaugural version of the AVS PCC Requirement was introduced, leading to the establishment of the AVS Working

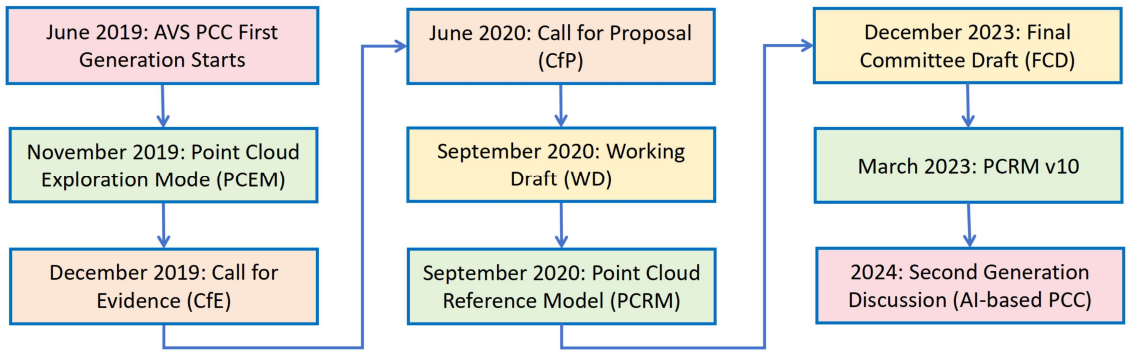

Figure 1: Milestones of the first generation AVS PCC standardization [7, 46, 50].

Group aimed at developing an independent point cloud compression standard in China. By November 2019, the first iteration of the Point Cloud Exploration Model (PCEM) was established as a foundational framework for gathering evidence regarding the feasibility of point cloud compression. Subsequently, in December 2019, the AVS PCC Call for Evidence (CfE) was published. In June 2020, a Call for Proposals (CfP) related to AVS PCC was issued [3]. Following this progression, September 2020 saw the release of the initial version of the AVS PCC Working Draft (WD), alongside which a preliminary version of Point Cloud Reference Model (PCRM) software was developed to provide a platform for exploring and assessing effective tools for point cloud compression [4]. The first edition of the AVS PCC Committee Draft (CD), along with its corresponding reference software PCRM v10, was released in February 2023 [6]. The finalization of the AVS PCC Final Committee Draft (FCD) is scheduled for December 2023 [5].

In terms of point cloud geometry coding, MPEG G-PCC employs a combination of predictive geometry coding and octree-based geometry coding [41]. It utilizes a predictive tree structure [36] to predict the positions of points and adaptively quantizes the azimuthal angle based on the radius, leading to improved compression performance. MPEG G-PCC also introduces inter-prediction techniques [45] that leverage global motion estimation for road and object points, enhancing the compression efficiency of point clouds captured by LiDAR sensors in moving vehicles. AVS PCC uses a reference software model known as Point Cloud Reference Software Model (PCRM) for geometric encoding. It includes algorithms for coordinate transformation, Morton code-based octree partitioning [100], and geometric coordinate prediction. AVS PCC's geometry compression is designed to handle the efficient encoding of point cloud data through various techniques such as block-based encoding and the use of a geometry data unit header for syntax modifications.

In terms of point cloud attribute coding, MPEG G-PCC's attribute coding involves techniques such as adaptive quantization [83] for attribute predicting transform coding, neighbor search methods for attribute Level of Detail (LoD) prediction [40, 57], and improvements to Region Adaptive Hierarchical Transform (RAHT) [17] attribute coding, including inter-prediction for DC and AC RAHT coefficients. AVS PCC's attribute compression includes methods for attribute preprocessing, prediction, quantization, and entropy coding. It utilizes techniques such as KD-Tree-based [8] resampling for attribute prediction and introduces the concept of attribute prediction using the geometric position of points. AVS PCC also discusses the use of color space transformations and the application of different prediction methods for various attribute types.

Each of these standards has been developed with the goal of achieving high compression ratios while maintaining the fidelity of the original point cloud data. They cater to different aspects of the compression process, achieving the optimization trade-off among complexity, efficiency, and the ability to reconstruct the point cloud accurately for various applications.

## 2 Fundamental Techniques of Point Cloud Compression

### 2.1 Overview

The basic procedure of the point cloud geometry coding technique is illustrated in Figure 2. Initially, the position information triples (x, y, z) for each point are collected or generated. While the data may be represented as floating-point numbers, n-bit integers are employed to encode the coordinates in accordance with the coding standard. The value of n determines the precision of the coordinates and necessitates their transformation. First, all point coordinates are normalized by subtracting their respective minimum values for x, y, and z components. Subsequently, quantization is performed to convert these normalized coordinates into corresponding integers. Second, based on the inherent characteristics of the data itself, MPEG G-PCC and AVS PCC can adopt distinct structures, namely octree and predictive tree, to effectively eliminate spatial redundancy within the original point cloud coordinates. Finally, information regarding either octree or predictive tree structure is entropy encoded into a geometry bitstream.

The basic route of point cloud attribute coding is also shown in Figure 2. First, the attribute information of the input point cloud will be transformed into a tractable space, for example, the RGB color attribute will be transformed into the tractable YUV space [72]. Then, due to potential reduction in the number of points during geometry compression, there may arise a mismatch between geometry and attribute information. Hence, the attribute values of the point cloud after geometry compression need to be recalculated. At this stage, pre-processed attribute information can be encoded using two meth-ods, i.e., multi-layer transformation-based coding and interpolation-based predictive coding. Both approaches may involve the processes of prediction and transformation, but the former performs multi-layer wavelet transformation on attribute information and processes all the attributes of the point cloud as a whole, while the latter directly predicts untransformed attributes of the point cloud. Finally, for both methods, quantization and entropy coding are applied to transform coefficients and prediction residuals.

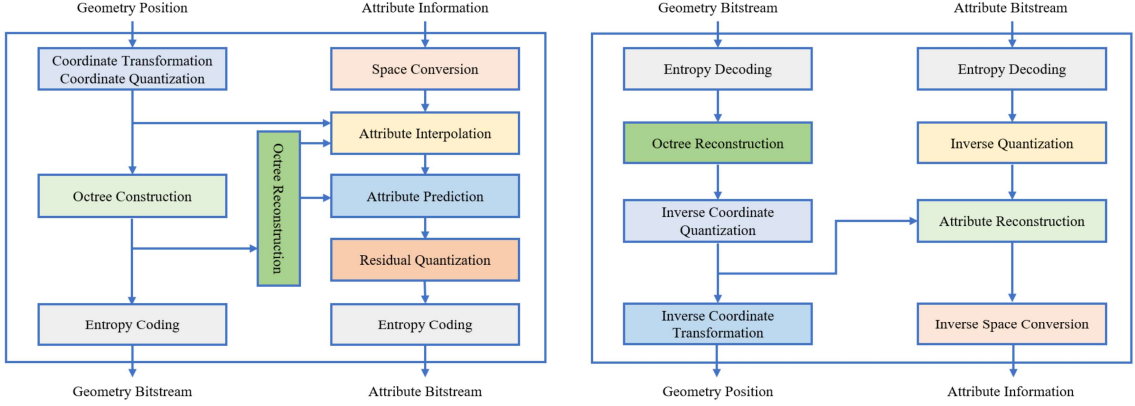

Figure 2: Flowchart of the general point cloud coding techniques [46, 48, 50].

### 2.2   *Point Cloud Geometry Coding*

- Octree coding

Octrees have been utilized for representing 3D objects since the 1980s [61]. Due to its simple structure and ability to efficiently handle local relationships within 3D information, octree is extensively employed in coding point cloud geometry. Assuming all points in the point cloud are enclosed within bounding boxes of size $D \times D \times D$, these bounding boxes are recursively divided into eight equal parts along the x, y, and z axes. For instance, after the initial division, the point cloud will consist of a maximum of eight bounding boxes measuring $\frac{D}{2} \times \frac{D}{2} \times \frac{D}{2}$ for each divided block. This process will continue iteratively until the size of bounding box reaches the minimum voxel unit size. During this procedure, whether the eight children of the current bounding box at each octree partition contain at least one point is recorded, and then it is denoted as 1 if it does, and 0 otherwise [46, 50]. The partitioning status of each node is represented by an 8-bit binary sequence known as an occupancy code. This occupancy code is subsequently sent to an entropy encoder for arithmetic coding.

Due to the varying size of the bounding box in different directions, denoted as $(2^{d_x}, 2^{d_y}, 2^{d_z})$, $d_x \neq d_y \neq d_z$, the octree partition cannot be extended indefinitely. Moreover, the point cloud data may exhibit sparse distribution characteristics along a specific direction, making quadtree and binary tree partitions more flexible for bounding box partition. The quadtree partition requires only a 4-bit occupancy code, while the binary tree partition needs only a 2-bit occupancy code. By employing an optimal partition method, better coding performance can be achieved [46, 50]. However, searching for this optimal method introduces significant time overhead. Different coding standards use adaptive algorithms to trade-off coding performance and time complexity. In cases where certain points are distant from the main part of the point cloud, these isolated points will cause the corresponding occupancy information to be 1 all the time, so that the bounding box is constantly partitioned in the whole partition process, leading to unnecessary coding overhead. Additionally, this portion's position information affects entropy coding context establishment and makes it challenging for context to accurately describe point distribution characteristics. Therefore, they are separately coded.

The geometry information of the point cloud is represented as an occupancy code, which is further entropy-encoded using a context-based binary arithmetic encoder [75]. The design of an effective context for utilizing the coded information becomes crucial in optimizing the octree coding algorithm. Additionally, different coding standards have also incorporated various algorithms to enhance coding performance based on specific characteristics exhibited by point clouds.

- Predictive tree coding

Octrees have demonstrated excellent performance in processing dense point clouds. However, as mentioned earlier, encoding geometry information in octrees necessitates a substantial number of points to construct intricate contexts, thereby requiring the encoding of a large number of points simultaneously. While real-time scenarios of point clouds, such as the application of point cloud in the field of autonomous driving, require low-latency processing data, and the number of points input in each encoding is limited, which is difficult to construct an effective context. As an alternative, a low-latency geometry encoding structure for point cloud -predictive tree is widely used [36]. The predictive tree is a tree structure, and each point in the point cloud is regarded as a node of the predictive tree. Each node selects the nearest node in the current tree as its parent node to join the predictive tree, so that every node except the root node can be predicted by its ancestor node, and the compression of geometry information is realized by encoding the prediction residuals. The accurate prediction of the current node and efficient encoding of residual information using ancestor nodes have emerged as key focal points in the algorithm for predictive tree coding [42].

### 2.3   *Point Cloud Attribute Coding*

- Multi-layer transformation-based coding

The basic process of point cloud attribute coding based on multi-layer transformation involves transforming the attribute information of the point cloud using wavelet transform to obtain direct current (DC) coefficients and alternating current (AC) coefficients [17]. Subsequently, the DC coefficients undergo successive wavelet transforms until reaching the final layer. All AC coefficients and the first layer's DC coefficients are quantized and subjected to entropy coding. In the course of the multi-layer transformation, the transform coefficients may be predicted, at which point the residuals of the transform coefficients will be encoded.

- Interpolation-based predictive coding

The fundamental element of interpolation-based predictive coding resides in directly predicting attribute information for data points. Initially, the point cloud undergoes grouping based on predetermined criteria, followed by applying interpolation-based prediction to each group while taking into account attribute information from adjacent points. Furthermore, a potential transformation can be conducted on the predicted residual. Eventually, the transformed coefficients or prediction residuals are entropy-coded [46, 48, 50].

### 2.4   *Typical Applications*

Use cases and applications related to AVS PCC, such as rate control [82, 98], are currently prevailing topics. We further elaborate on the related developments. [97] designs a rate control scheme for AVS-PCC-PCRM about LiDAR point cloud sequence, and experiments on Cat2-frame sequence achieve good rate control effect, and BD-GeomRate and BD-AttrRate have almost no effect. In [96], an attribute bit control scheme for AVS-PCC-PCRM is proposed, and experiments are carried out on Cat2-frame sequences, which achieves good rate control effect and has a certain degree of bit fluctuation stabilization effect.

## 3   AVS Point Cloud Compression Techniques and Standard

### 3.1   *Octree Coding for Geometry Compression in AVS PCC*

AVS PCC adopts implicit partition method for bounding box partition. Considering the trade-off between coding performance and complexity, the geometry partition method of each layer is the same, and two hyperparameters $K$ and $M$ are used to control it, where $K$ means binary tree or quadtree partition can be used for the first $K$ layers, and M means octree partition is used for the last $M$ layers [122]. There are two approaches for geometry partition, where the first is to force the partition in all three directions, and the second is to make the bounding box sizes of the three directions tend to be the same by partitioning. Specifically, when there is a direction with the largest bounding box size, the binary tree partition is carried out in that direction, and when there are two directions with the same bounding box size and larger than the third direction, the quadtree partition is carried out along the two directions. The octree partition is performed when the sizes of the three directions are the same. The geometry partitioning process is divided into four stages according to $K$ and $M$. The first stage is the first $K$ layers, at which time the second logic is used for partitioning. The second stage is to divide continuously until the size of the bounding box in one direction is equal to $M$, at which point the first logical division is adopted. The third stage is to continue dividing until there are $M$ layers left, at which point the second logical division is adopted. The fourth stage is the last $M$ layers, at which point the first logical division is adopted.

AVS PCC incorporates three constraints for handling isolated points, including the user-defined isolated point coding mode can be activated, the current layer permits enabling of the isolated point coding mode, and the current node contains only one point. In such cases, the division of the node will stop, occupancy information corresponding to that node becomes 0 and the isolated point is encoded separately. In the process of dividing the bounding box, whether the isolated point coding mode is allowed for different layers is judged according to whether the ratio of the number of points in the current layer to the number of nodes in the current layer is less than the set threshold [126, 46].

The resulting occupancy code is fed into a context-based adaptive binary arithmetic encoder for encoding. When the partition of the current block is encoded, the occupancy of all blocks in the same layer of the current block and the occupancy of all blocks that have been encoded before the block to be encoded according to the coding order are known. The former is called the same layer information and the latter is called the sub-layer information. AVS PCC designs the corresponding sets of contexts for sparse point clouds and dense point clouds according to the occupancy of adjacent blocks of the block to be encoded. The two sets of contexts are composed of the occupancy of adjacent blocks at different positions [123, 125, 46]. The sub-layer information of the first set of context suitable for sparse point cloud includes 3 sub-blocks in the current block to be encoded coplanar, 3 co-edges, 1 co-point, 1 sub-block in the shortest direction of the block to be encoded (in the opposite direction of the coding order, and 2 current sub-layers of the corresponding length from the current sub-block to be encoded), and the 5 coded sub-blocks in the partitioned coplanar and collinear adjacent blocks of the current block that are at the same position as the sub-block to be coded. The same layer information includes the occupancy of the three parent blocks of the same layer which are coplanar with the child block to be encoded. The sub-layer information of the context suitable for dense point cloud includes 3 sub-blocks in the current block to be encoded coplanar, 3 co-edges, 1 co-point, and the coded sub-block which is the same as the sub-block to be coded in the position of the current block among the 3 adjacent blocks which have been divided and coplanar with the current block. The same layer information includes the occupancy of the 6 parent blocks of the same layer that are directly adjacent to the child block to be encoded, and the 6 blocks of the same layer that are coplanar with the current block [125].

AVS PCC also designs the planar coding mode to deal with the case that the points are almost only distributed in half plane. When the planar coding mode is turned on, before encoding the current block for the slice with a smaller ratio of bounding box and number of points, it checks whether the five coded neighbor nodes in the same layer as the reference block have more than three children nodes and half plane is not occupied. If the condition is satisfied, it enters the planar coding mode, and designs a new context according to whether the sub-block occupancy of the reference block satisfies the planar state and the plane in which the sub-block to be encoded is located [123, 46].

### 3.2 Comparison with Octree Coding in MPEG G-PCC

MPEG G-PCC cannot only adopt the same implicit geometry partition as AVS PCC, but also express the partition explicitly through 3-bit syntax elements [48, 50]. These three bits indicate whether the x, y, and z axes are divided or not. The explicit partitioning approach provides more flexible control over the bounding box partition [79].

MPEG G-PCC designs three isolated point coding modes with different relaxed degrees, controls the entry of isolated point coding mode based on the occupancy information of parent node and adjacent blocks, and performs isolated point coding when the number of points in a block is less than or equal to 2. Compared with AVS PCC [46, 50], MPEG G-PCC is more flexible in determining isolated point [48]. The partitioned occupancy codes can also be fed into the context-based adaptive binary arithmetic encoder for encoding. The context in MPEG G-PCC [48] also contains the occupancy information of neighboring blocks, and together with the occupancy information of six neighboring blocks that are coplanar with the parent block of the current child block to be encoded, forms the earliest neighborhood context. One bit is used to represent the occupancy of each position, there are 64 contexts. In addition, when encoding 8-bit occupancy code, the encoded information is used as the context of the remaining unencoded bits after encoding each bit.

The neighborhood contexts can be reduced to 10 by the similarity of point cloud geometry information. According to the position of the sub-block to be encoded, part of the neighborhood occupancy information of the post-encoded sub-block may represent the same information as the context composed of encoded bits, therefore different numbers of neighborhood contexts can be used for different positions of the block to be encoded to further reduce the number of neighborhood contexts. Besides, the same sub-block occupying case that has encoded other neighborhood blocks as AVS PCC is introduced [124].

In addition to the occupancy information of adjacent blocks, MPEG G-PCC also designs other aspects of context. For example, the position and occupancy of the 26 adjacent blocks adjacent to the current block are used to score the occupancy of the sub-block to be encoded [48]. The scores are divided into three outcomes based on two set thresholds: inability to make predictions, prediction of occupancy, and prediction of non-occupancy. The 26 adjacent blocks can be further reduced. The context based on prediction and the context based on neighborhood block occupancy information include the occupancy information of basic six neighborhood block, the encoded bit information of the current block, and the encoded sub-block occupancy information of neighboring blocks to formulate a new context [60].

MPEG G-PCC also has planar coding mode, but unlike AVS PCC, which designs the context according to the situation that the planar mode of the reference block and the subblock to be encoded are satisfied, MPEG G-PCC encodes the planar coding mode related information explicitly. According to the density of points in the current block and the probability that the encoded block satisfies the planar coding mode, it decides whether to enter the planar coding mode. If it enters, a 1-bit flag is encoded explicitly to indicate whether the plane condition is satisfied. If the flag is 1, another 1-bit flag is encoded to indicate that the point is located in the upper or lower half plane. The planar coding mode can be applied in up to three directions, and the partial occupancy information of the current block can be directly determined based on this information. MPEG G-PCC can also use the prior information of the points obtained from the rotating LiDAR point cloud to assist the coding process of planar coding and direct coding, which is called angle coding mode [43].

### 3.3 Predictive Tree Coding for Geometry Compression in AVS PCC

The predictive tree in AVS PCC is constructed as a linear linked list [27, 46]. Each node, except the root node, has a unique parent node, and all the points that have not yet joined the predictive tree form a KD-tree. Each time, the nearest neighbor point of the current predictive tree leaf node is searched in the KD-tree as the child node of the leaf node to join the predictive tree, and the nearest neighbor point is moved out of the KD-tree. This process continues until all points within the KD-tree are added to the predictive tree. Each non-root node undergoes direct prediction by its parent node, with geometry information being represented through encoded residuals. The residual encoding procedure comprises two steps, i.e., residual sign encoding based on geometry information and absolute value encoding based on context [130, 46].

### 3.4  Comparison with Predictive Tree Coding in MPEG G-PCC

The predictive tree structure of MPEG G-PCC exhibits greater complexity and flexibility for various application scenarios. MPEG G-PCC uses a ternary tree [18] to build a predictive tree [64], each node can have a maximum of three children. Unlike AVS PCC, when constructing the predictive tree, MPEG G-PCC identifies the point with the closest distance to the current point and fewer than three child nodes as the parent node for the current point based on application scenario constraints. According to the required encoding speed and encoding delay of the application scenario, the construction of the predictive tree will change. Under high-delay fast mode, all points are sorted using Morton order, and each search considers a certain number of neighbor points obtained through Morton order as the range for searching nearest neighbor points. In the high-delay slow mode, nearest neighbor search is performed by traversing all points based on the KD-tree. In the low-delay mode, a buffer is maintained where points are processed sequentially based on their input, and the nearest neighbor point search is performed within this buffer range. Additionally, in MPEG G-PCC, the point is also predicted by ancestor nodes, and up to three ancestor nodes are used simultaneously. There exist four patterns as follows: utilizing the default value as the predicted value; employing the parent node as the predicted value; incorporating both parent and grandfather nodes for prediction; integrating parent, grandfather, and great-grandfather nodes for prediction [24].

MPEG G-PCC traverses the predictive tree by depth first search, employing rate distortion optimization to select the optimal coding mode for each node to be encoded and recording the selected mode [48]. Additionally, the prediction residual is encoded. Moreover, a specialized predictive tree structure is proposed for the coefficient point cloud obtained from rotating LiDAR with known parameters. This novel predictive tree fully leverages the characteristics of point cloud data collected by rotating LiDAR and transforms Cartesian coordinates into an $(r, \phi, i)$ coordinate system that better describes LiDAR point clouds. Herein, r represents radius while $\phi$ denotes rotation angle, i corresponds to the serial number of the respective radar head inclined at a specific angle. The point cloud is acquired through by a series of lasers rotating around the vertical direction at a certain speed with different inclination angles [64]. For each point in Cartesian coordinate system, the radar head serial number can be uniquely determined by the different heights of radar heads. In the new coordinate system, the construction process of the predictive tree dispenes with the computationally complex nearest neighbor search process. Points with the same i form a linked list structure. If a linked list exists for a given current point, it will be directly added. Otherwise, if no linked list exists for said current point, its parent node will be selected from neighboring points belonging to other linked lists. This way the predictive tree is built with very low complexity. Due to the full use of rotating Li-DAR information, the proposed method can achieve similar geometry coding performance as the predictive tree based on Cartesian coordinate system [23].

### 3.5 Multi-layer Transformation-based Coding for Attribute Compression in AVS PCC

Based on multi-layer transformation, the point cloud attribute coding in AVS PCC [46] constructs a hierarchical structure [11]. All points are sorted according to the Hilbert code of their geometry coordinates to form the bottom layer. In each layer, points can either serve as prediction points or transformation points. If the distance between two points is below a predefined threshold, they are transformation points and merged into a new point in the upper layer. Otherwise, the first point becomes the prediction point while the second point serves as the current point for further iterations. After all the points in the bottom layer are processed, the upper layer is traversed. When there are fewer than 128 points or more than half of all points remaining in a subsequent layer, all points within that layer are directly paired and merged together. This process continues until reaching the first layer with only one point remaining. The geometry coordinates of all parent nodes represent an average value calculated from their respective children's coordinates, while the distance threshold is dynamically updated throughout this construction process [12].

After constructing the multi-layer structure, the DC and AC coefficients are obtained through perform wavelet transform for transformation points from bottom to top. The predicted value of the prediction point is derived by taking a weighted average of the reconstructed DC coefficient from three neighboring points within the same layer, while also calculating the prediction residual. The final DC coefficients of the first layer, all AC coefficients and prediction residuals are entropy coded. The weight used for predicting can be determined based on either Manhattan distance between two points or a combined measure considering both geometry and attribute information in point clouds with multiple attribute categories [11]. An illustration of nearest points and equidistant points in Manhattan distance.

### 3.6 Comparison with Multi-layer Transformation-based Coding in MPEG G-PCC

MPEG G-PCC employs region-adaptive hierarchical transformation algorithm (RAHT). In this approach, each occupied parent block is divided into octrees based on their inherent structure. A transformation is applied to each parent block that contains eight sub-blocks, starting from the bottom and progressing upwards. Notably, these transformations occur exclusively between two occupied sub-blocks. Each parent block is transformed along the three coordinate axes in turn, and finally a DC coefficient and several AC coefficients are obtained. The AC coefficients are quantized and entropy coded, and the DC coefficients participate in the transformation process of the next layer [17].

To further improve the coding performance, RAHT introduces inter-layer prediction, which changes the bottom-up coding order to top-down, and the transformation is still performed within the parent block containing at most eight child blocks. The reconstructed attribute values of the parent node and the neighbor nodes of the parent node are used to predict the attributes of the child nodes in the parent node, and the attribute predicted values of the child nodes are obtained. Then the RAHT transformation is applied to the true attribute values and the predicted attribute values of the child nodes to obtain the corresponding DC coefficients and AC coefficients. The AC coefficients obtained by the real attribute value transformed and the high frequency AC coefficients obtained by the predicted attribute value transformed are subtracted to obtain the AC coefficient residuals, which are quantized and entropy coded [34].

The DC coefficient corresponding to each block to be transformed is the attribute in the block divided by $\sqrt{w}$. The prediction of the attributes is performed in the attribute mean domain, so the conversion from the DC coefficients to the attribute mean within the block is performed. After the attribute mean of all sub-blocks is predicted, the transformation from attribute mean to DC coefficient is carried out. After that, RAHT transform is performed on the real DC coefficients and predicted DC coefficients of each sub-block respectively, and the corresponding AC coefficients are subtracted. The obtained AC coefficients are quantized and entropy coded. The DC coefficients are passed continuously during the transformation, and only the DC coefficients of the first layer need to be encoded.

### 3.7 Interpolation-based Coding for Attribute Compression in AVS PCC

Interpolation-based attribute prediction coding algorithm in AVS PCC is generally utilizing the geometry distance of point cloud to predict the attribute [46, 50]. Point cloud represents a common object or scene, and its adjacent points have similar attribute values. Using similar points in spatial distance to predict the attribute of the current point is one of the common algorithms.

First, attribute sorting sorts the geometry position of the point cloud after geometry compression. The common sorting schemes are based on Morton order and Hilbert order [16]. Based on the points of Morton attribute, the Morton code of the current point is first found, and then sorted according to the size of the Morton code. In AVS PCC, Morton code is obtained by looking up the table. If Morton code is calculated in the encoding, the time complexity will be greatly increased, and the time can be greatly saved by looking up the table. The disadvantage of Morton order is that the geometry position of similar Morton code points will appear jump phenomenon. Similarly, point cloud sorting based on Hilbert order can solve the jump phenomenon to a certain extent. When using the method based on Hilbert ordering and calculating the Hilbert order, it is necessary to introduce a bias coefficient $\theta$ to transform the coordinates of the point cloud $X = (x, y, z)$ into $X^\theta = (x, y, \theta \times z)$. The transformed coordinates are used to iteratively query the table to obtain the Hilbert code of the current point, and finally the points are sorted according to the Hilbert order. The two kinds of sorting are different by the data set, that is, in the compressed configuration file, different data sets may be sorted differently [94].

For the sorted point cloud, it needs to select and predict its neighbor points. One of the methods is the color attribute prediction method based on double Morton order. Because Morton order has the property of jumping, in order to solve this problem, the double Morton order-based color attribute prediction is introduced. The coordinates of the point cloud are first obtained and Morton codes for all point sets are generated to obtain Morton order one. For the current point to be encoded, add the $M$ points that have been encoded before the current point to the cache, where $M$ is a parameter set at the encoder, which needs to be encoded at the encoder. For the points in the cache, add a fixed offset coefficient $C$ to the coordinates of each point, that is, calculate the Morton code for the new point in the cache. Morton sorting is performed to obtain Morton order two [46, 50].

Level of detail (LoD) color attribute prediction based on Hilbert order first performs a fine hierarchical operation on the point cloud, which is called the construction of LoD structure [46, 50]. The prediction operation is performed on the LoD of each layer. The LoD hierarchical coefficient $K$ is first determined, which represents the number of right shifts of the initial Morton code. That is, within the initial search range $2K$, its sampled neighbor points are not less than 0.6. Suppose that the current point cloud can be divided into $N$ layers, and each layer is a LoD. Firstly, all the points of the point cloud are added to the $Nth$ level of LoD, and each point in the $Nth$ level of LoD is traversed. If the current point is not visited, it is divided into the $N - 1$ level of LoD. The neighbor points of the common edge are divided into the points in the middle layer of level $N$ and level $N - 1$ of LoD and marked as visited. The next point is traversed and the process is carried out iteratively. For the next level of LoD, the size of the iteration block is increased to the parent block. After different levels of LoD division, for the $Nth$ level of LoD, the points of the $N-1$ level of loD will be selected as prediction points and added to the prediction point set, and the points in the middle level will be predicted. If the intra-layer prediction technology is enabled, the points encoded earlier than the current point in the same LoD layer will also be added to the prediction point set. Finally, the attribute of the current point is weighted by the points in the prediction point set [130].

In AVS PCC, the reference point set is updated as follows. If the number of points in the reference point set is less than the maximum number of nearest neighbors specified at that time, the reconstructed points are inserted into the reference point set. However, if the number of points in the reference point set equals the maximum number of nearest neighbors, then we identify and replace with a reconstructed point, the one in the reference point set that is farthest from the current point [131, 46].

The compression performance of point clouds with multiple attribute types can be enhanced through cross-attribute prediction. The calculation of the comprehensive distance involves determining the geometry distance between the point to be encoded and its neighbors, calculating the attribute distance based on the absolute value of the residual between the attribute type of the point to be encoded and its neighbors, and combining both geometry and attribute distances in a comprehensive manner.

### 3.8  *Comparison with Interpolation-based Coding in MPEG G-PCC*

MPEG G-PCC also employs a predictive coding technique based on the LoD. Firstly, the LoD is constructed, and subsequently, unlike AVS PCC, the attribute predicted value of the point to be encoded is estimated through nearest neighbor search and weighted average calculation using the points contained in either the previous or current level's LoD. Finally, this prediction residual is encoded.

The LoD structure is a hierarchical structure, which corresponds to the point cloud with gradually smaller down-sampling degree from bottom to top. Initially, the original point cloud is sorted based on Morton code. To meet low-latency attribute encoding requirements, only one level of detail can be utilized without sorting at this stage. Subsequently, the point cloud is divided into different levels of detail using a series of distance thresholds. When building the ith layer, the sorted points that has not yet been added to the level of detail is traversed, and a new middle layer is established. Except for the first point that is directly added to the middle layer, every time the distance between the current point and the last added middle layer point is greater than the threshold of the layer, the current point is added to the middle layer. The distance threshold for each level progressively decreases until all points are incorporated into their respective levels of detail [78, 90].

After LoD is established, it starts from the highest LoD to encode some points in two adjacent layers which only belong to the higher level. For each point to be encoded, $K$ neighbors are searched by nearest neighbor search based on the distance between the two points in the lower LoD. If the same level prediction is allowed, the search range includes all the points encoded before the current point in the same level. Then, $K$ prediction points are selected to predict separately according to the rate-distortion optimization or the weighted average of the attributes is calculated based on the distance between two points as the predicted value. Filtering the weights based on distance according to the position distribution of the selected reference points can further improve the coding performance [89].

In lossy coding, MPEG G-PCC incorporates dynamic weight updating and adaptive quantization processes into the aforementioned algorithm. The more times the reference points are selected, the greater the prediction weight is. During quantization, the distortion degree is changed according to the prediction weight, so that the reconstruction quality of the points that are selected for many times is higher, and the accuracy of the prediction is improved.

### 3.9   Summary

The main technical aspects of AVS PCC and MPEG G-PCC can be summarized as depicted in Figure 3. In terms of technicality, both AVS PCC and MPEG G-PCC exhibit similar branches. However, notable disparities exist in their specific implementations.

## 4   Comparative Analysis of Coding Performance

### 4.1   Comparison with Learning-based Standard: JPEG AI PCC

#### 4.1.1   Goal and Purpose

AVS PCC and JPEG AI PCC have distinct goals and applications: AVS PCC primarily focuses on providing efficient point cloud compression, suitable for applications requiring precise 3D geometric representations such as 3D modeling and virtual reality [46]; whereas JPEG AI PCC utilizes deep learning technology, aiming to provide a unified compressed representation for both human vision and machine processing, suitable for applications that require direct processing within the compressed domain, such as point cloud classification and detection, marking a new direction in multimedia compression technology [2, 35].

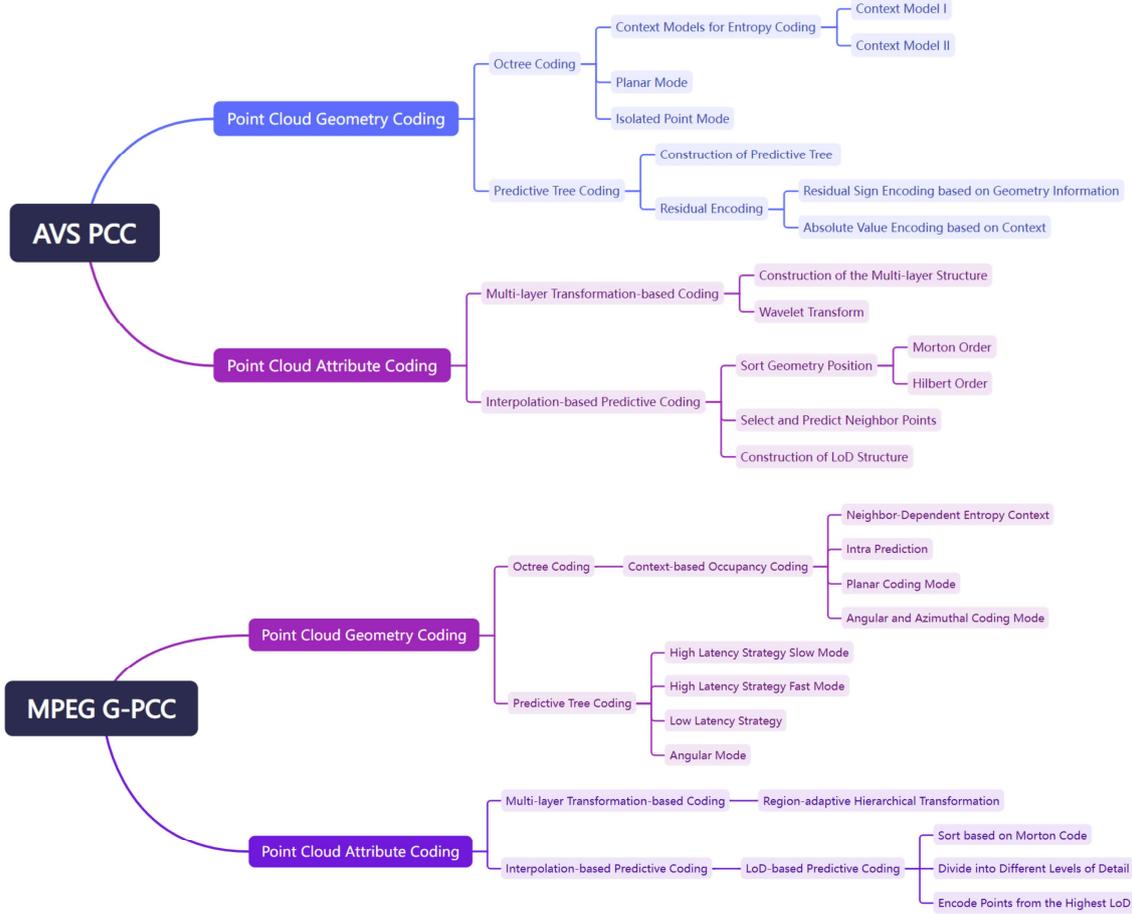

Figure 3: Critical techniques in AVS PCC and MPEG G-PCC.

### 4.1.2   *Performance and Complexity*

AVS PCC and JPEG AI PCC show significant differences in encoding methods, performance, and complexity: AVS PCC mainly adopts traditional geometry-based encoding techniques, focusing on direct processing of the 3D geometric information of point clouds, and optimizes encoding through prediction and transformation techniques such as LoD and RDO strategies, while geometry and attribute data are usually processed separately [27]. In contrast, JPEG AI PCC is based on deep learning, using sparse convolutional neural networks to directly process raw 3D geometric data, and projects color data onto 2D images for encoding, achieving joint encoding of geometry and color data. In terms of performance, AVS PCC excels in geometric encoding, while JPEG AI PCC achieves significant bit rate reduction in geometric encoding, although its color compression performance is slightly inferior, its learning framework improves overall performance. In terms of complexity [2, 35], AVS PCC has high encoding complexity when performing LoD construction and fine geomet-ric processing, and JPEG AI PCC also has high encoding complexity due to the use of deep learning models during training and inference stages, but it reduces the computational complexity of processing and computer vision tasks through learning models, as it skips the actual decoding steps and subsequent feature extraction. These differences reflect the trade-offs and optimizations of different performance indicators in the design of the two standards.

### 4.2 Comparison with Non-learning-based Standard: MPEG G-PCC

It should be noted that MPEG V-PCC [49] uses projection and video coding to implement the point cloud compression task, while AVS PCC [46] and MPEG G-PCC [48] adopt different compression strategies from the point cloud geometry domain. Therefore, we compare these two standards in terms of both coding efficiency and coding time. The test software versions are MPEG G-PCC's TMC13 v22 and AVS PCC's PCRM v13. The hardware platform is Intel© Core™ i5 10500.

#### 4.2.1 Comparison of Geometry Coding Efficiency

We compare the results of lossless geometric encoding and lossy geometric encoding separately. When performing lossy geometric encoding, in order to obtain different bit rate points, the parameter *positionQuantizationScale* is set as 0.125, 0.25, 0.75, 0.875 and 0.9375, respectively.

- Octree coding

Table 1 shows the bitrate gain of octree coding in AVS PCC compared with octree coding in MPEG G-PCC. It can be observed that the coding performance of AVS PCC is better than MPEG G-PCC in octree-based geometry coding [119]. For the lossy and loseless coding modes, compared with the MPEG G-PCC, the AVS PCC can have -7.92% and -10.33% bitrate reductions, respectively.

Table 1: Geometry bitrate gain of octree coding in AVS PCC compared with octree coding in MPEG G-PCC.

| Point Clouds | Lossy (%) | Lossless (%) |
|---|---|---|
| basketball_player_vox11_00000200 | -10.10 | -13.03 |
| dancer_vox11_00000001 | -8.70 | -11.75 |
| longdress_vox10_1300 | -8.40 | -9.70 |
| loot_vox10_1200 | -7.20 | -9.83 |
| redandblack_vox10_1550 | -6.50 | -9.29 |
| soldier_vox10_0690 | -6.60 | -8.40 |
| Average | -7.92 | -10.33 |

● Predictive tree coding

The results of geometry lossless coding performance between prediction tree in AVS PCC and MPEG G-PCC are shown in Table 2. It can be seen that AVS PCC is better than MPEG G-PCC [33]. AVS PCC can save 14.96% bitrate compared with MPEG G-PCC in geometry lossless coding.

Table 2: Comparison of geometry lossless coding performance between prediction tree in AVS PCC and MPEG G-PCC.

| Point Clouds | AVS PCC (bpp) | MPEG G-PCC (bpp) | Bitrate Gain (%) |
|---|---|---|---|
| citytunnel | **15.79** | 17.74 | -10.99 |
| overpass | **18.02** | 20.77 | -13.21 |
| tollbooth | **16.26** | 20.50 | -20.67 |
| Average | **16.69** | 19.67 | -14.96 |

### 4.2.2   Comparison of Attribute Coding Efficiency

Similarly, we compare the results of lossless attribute encoding and lossy geometric attribute encoding based on lossless geometry. When performing lossy geometric encoding, in order to obtain different bit rate points, the parameter $qp$ is set as 22, 28, 34, 40, 46, and 51 respectively.

● Multi-layer transformation-based coding

Bitrate gain of multi-layer transformation-based coding in AVS PCC compared with multi-layer transformation-based coding in MPEG G-PCC is shown in Table 3. It can be seen that AVS PCC is worse than MPEG G-PCC [13]. AVS PCC performs is inferior to MPEG G-PCC in multi-layer transformation-based coding by consuming an average of over 50% more bitrates.

Table 3: Attribute bitrate gain of multi-layer transformation-based coding in AVS PCC compared with multi-layer transformation-based coding in MPEG G-PCC.

| Point Clouds | Lossy (%) | Lossless (%) |
|---|---|---|
| basketball_player_vox11_00000200 | 76.30 | 79.80 |
| dancer_vox11_00000001 | 75.30 | 75.50 |
| longdress_vox10_1300 | 34.50 | 58.90 |
| loot_vox10_1200 | 50.70 | 124.70 |
| redandblack_vox10_1550 | 52.00 | 77.10 |
| soldier_vox10_0690 | 51.30 | 113.30 |
| Average | 56.68 | 88.22 |

● Interpolation-based predictive coding

Bitrate gain of interpolation-based predictive lossless coding in AVS PCC compared with MPEG G-PCC is shown in Table 4. AVS PCC performs better compared to Table 3, but still slightly worse than MPEG G-PCC in interpolation-based predictive lossless coding [59], where the average 5.34% bitrate increase can be observed.

Table 4: Attribute bitrate gain of interpolation-based predictive lossless coding in AVS PCC compared with MPEG G-PCC.

| Point Clouds | Bitrate Gain (%) |
|---|---|
| basketball_player_vox11_00000200 | -1.06 |
| dancer_vox11_00000001 | 9.18 |
| longdress_vox10_1300 | 5.76 |
| loot_vox10_1200 | 6.08 |
| redandblack_vox10_1550 | 7.56 |
| soldier_vox10_0690 | 4.50 |
| Average | 5.34 |

### 4.2.3   Comparison of Computation Time

We have participated into the experiments for computation time comparison between AVS PCC and MPEG G-PCC, and the results can be achieved from the extensive experiments as shown in the corresponding AVS proposals [13, 117, 59].

In the experiment comparing encoding complexity, the datasets used are CAT1A [65] and CAT2 [66]. Geometry was encoded using lossless compression, while attributes were encoded using lossy compression with $qp = 8$.

For Transformation-based coding, the encoding and decoding times of AVS PCC attribute compression are similar to those of MPEG G-PCC, but the encoding and decoding times of AVS PCC geometry compression are larger than those of MPEG G-PCC when testing transformation-based coding [13, 117].

For Prediction-based coding, the encoding and decoding time values of AVS PCC attribute compression are slightly shorter than those of MPEG G-PCC, but the encoding and decoding time values of AVS PCC geometry compression are almost twice those of MPEG G-PCC when testing prediction-based coding [59].

### 4.2.4   Summary

In terms of coding efficiency, the encoding performance differences between AVS PCC and MPEG G-PCC in point cloud geometry and attribute encoding can be attributed to their distinct approaches to prediction and transformation algorithms, entropy encoding methods, and quantization processes. Each standard's design accommodates particular application scenarios and performance requirements, leading to variations in encoding performance. These differences also reflect the trade-offs and optimizations made during the design of the two standards for different performance indicators.

AVS PCC and MPEG G-PCC show varying performance in point cloud geometry encoding. AVS PCC utilizes different geometric encoding methods, which sometimes result in better performance, while in other cases, MPEG G-PCC outperforms due to its complex rate-distortion optimization strategies that are effective for complex geometric structures [59].

In attribute encoding, AVS PCC often demonstrates superior performance, particularly in transformation attributes, likely due to its efficient prediction and transformation algorithms that are effective for specific attribute distributions within point cloud data [13]. MPEG G-PCC, on the other hand, uses methods such as wavelet-based attribute compression and LoD prediction, which show better performance in certain cases but not in others.

The differences in encoding performance between AVS PCC and MPEG G-PCC stem from their choices in prediction and transformation algorithms, entropy encoding methods, and quantization processes. Each standard's design considers specific application scenarios and performance requirements, leading to variations in encoding performance. These differences also reflect the trade-offs and optimizations made during the design of the two standards for different performance indicators.

In terms of coding time, as for geometry encoding, AVS PCC employs a prediction plus transformation algorithm that includes Hilbert sorting and neighbor searching, as well as adaptive transformation, while MPEG G-PCC utilizes a Lifting algorithm that involves attribute prediction based on LoD and lifting transformation. AVS PCC takes longer in the LoD and neighbor searching part because it uses a more complex Hilbert code, while G-PCC takes longer in the transformation part because it requires quantization weight calculation and lifting transformation of predicted residuals based on LoD [119]. In point cloud attribute encoding, AVS PCC's prediction algorithm encoding and decoding time is longer than G-PCC's because it directly differentials the encoded values and predicted values, while G-PCC performs weighted prediction between LoD layers. G-PCC's attribute entropy encoding (and decoding) time complexity is higher than AVS PCC's, possibly due to the inherent complexity of G-PCC's entropy encoding process and its point-by-point encoding method [117, 33],. These differences mainly stem from their different choices in prediction and transformation algorithms, entropy encoding methods, and quantization processes.

## 5 Conclusion

This paper provides a thorough review on the recently developed AVS point cloud compression standard, including the involved technologies and the per-formance evaluations. With the continuous advancement of technologies, point cloud compression is increasingly pivotal in enhancing the storage efficiency of point cloud data and reducing transmission cost. The first generation AVS point cloud compression standard has come to its final stage after submitting the final committee draft. It should be noted that, as the fast growth of deep learning-based approaches [47], the AVS workgroup also started the discussion for the end-to-end learning-based point cloud compression technologies and standards in December 2023 [32]. We can see that numerous challenges and issues still necessitate further explorations in point cloud compression. Furthermore, we can also see the first work to utilize large language model for point cloud compression [112]. Hence, we anticipate an increasing number of researchers devoting themselves to this research domain in the future, collaborating to propel the technologies and associated standard activities of 3D point cloud compression.

## Acknowledgements

This work was supported by The Major Key Project of PCL (PCL2024A02), Natural Science Foundation of China (62271013, 62031013), Guangdong Provincial Key Laboratory of Ultra High Definition Immersive Media Technology (2024B1212010006), Guangdong Province Pearl River Talent Program (2021QN020708), Guangdong Basic and Applied Basic Research Foundation (2024A1515010155), Shenzhen Science and Technology Program (JCYJ20240813160202004, JCYJ20230807120808017), Shenzhen Fundamental Research Program (GXWD20201231165807007-20200806163656003).